\documentclass{article}
\usepackage{spconf,amsmath,graphicx}
\makeatletter
\def\section{\@startsection{section}{1}{\z@}%
  {-0.9ex plus -0.2ex minus -0.2ex}%
  {0.4ex plus 0.2ex}%
  {\large\bfseries}}
\def\subsection{\@startsection{subsection}{2}{\z@}%
  {-0.8ex plus -0.2ex minus -0.2ex}%
  {0.3ex plus 0.2ex}%
  {\normalsize\bfseries}}
\makeatother

\usepackage{enumitem}
\setlist{nosep, leftmargin=14pt}

\usepackage{mwe} 
\usepackage{multicol}
\usepackage{multirow}
\usepackage{xcolor}

\title{Vessel-Aware Deep Learning for OCTA-Based Detection of AMD}
\name{Margalit G. Mitzner, Moinak Bhattacharya, Zhilin Zou, Chao Chen, Prateek Prasanna}
\address{Stony Brook University, Stony Brook, NY, USA}
\begin{document}
%
\maketitle

\begin{abstract}
Age-related macular degeneration (AMD) is characterized by early micro-vascular alterations that can be captured non-invasively using optical coherence tomography angiography (OCTA), yet most deep learning (DL) models rely on global features and fail to exploit clinically meaningful vascular biomarkers. We introduce an external multiplicative attention framework that incorporates vessel-specific tortuosity maps and vasculature dropout maps derived from arteries, veins, and capillaries. These biomarker maps are generated from vessel segmentations and smoothed across multiple spatial scales to highlight coherent patterns of vascular remodeling and capillary rarefaction.
Tortuosity reflects abnormalities in vessel geometry linked to impaired auto-regulation, while dropout maps capture localized perfusion deficits that precede structural retinal damage. The maps are fused with the OCTA projection to guide a deep classifier toward physiologically relevant regions. Arterial tortuosity provided the most consistent discriminative value, while capillary dropout maps performed best among density-based variants, especially at larger smoothing scales. Our proposed method offers interpretable insights aligned with known AMD pathophysiology. 

\end{abstract}

\section{Introduction}
Age-related macular degeneration (AMD) is a leading cause of irreversible vision loss worldwide, highlighting the importance of quantitative biomarkers for early detection and intervention~\cite{lad2023biomarkers,de2015review,told2016comparative}. OCTA provides high-resolution, non-invasive visualization of the retinal microvasculature and has become a valuable modality for biomarker discovery~\cite{hormel2023oct,told2023octa,hsieh2019oct,lu2024computational}. Machine learning approaches have been applied to OCTA for AMD analysis, with early work relying on hand-crafted texture descriptors classified using traditional ML models~\cite{alfahaid2025machine}. More recent DL methods, including Convolutional Neural Networks (CNNs) trained on OCTA images, have demonstrated strong performance in grading AMD activity~\cite{ran2021deep,yan2020deep,lee2017deep,yim2020predicting,peng2019deepseenet}. However, despite their predictive power, these models primarily exploit global image appearance and overlook rich clinical priors related to vascular structure—leaving them wanting in interpretability and biological alignment.
Although OCTA-derived metrics such as reduced capillary density have been proposed as potential early markers of AMD~\cite{dugiello2024vascular,mcleod2009relationship,bhutto2012understanding}, vessel-type–specific properties, including density and tortuosity across distinct vessel types, remain largely unexplored. Existing DL methods rarely incorporate vessel-level priors that correspond to clinically meaningful patterns of vascular dysfunction, limiting their ability to provide mechanistic insight. In contrast, OCTA analyses in diabetic retinopathy and other vasculopathies routinely segment and quantify microvascular features such as capillary dropout, non-perfusion areas, FAZ geometry, and vessel tortuosity, all of which correlate strongly with disease severity~\cite{moir2021review,sun2021optical}. Similar vessel-level biomarkers have been described in retinal vein occlusion. Yet, despite these advances, explicit computation and integration of vessel-level biomarkers for AMD remain limited.

\noindent\textbf{Contributions}. To address this gap, we propose a framework that integrates vessel-specific density and tortuosity metrics into a deep classifier via multiplicative attention. This approach allows for assessment of how vascular features across arteries, veins, and capillaries contribute to AMD discrimination. We also compare vessel-specific attention across increasing smoothing scales to determine which vessel type and feature provide the most consistent discrimination, providing new insight into early microvascular signatures of AMD. Extensive experiments demonstrate that incorporating vessel-level priors enhances interpretability while maintaining competitive performance, highlighting the value of biologically-informed modeling for retinal disease classification.

\section{Methodology}
\begin{figure}[t]
    \centering
    \includegraphics[width=\linewidth]{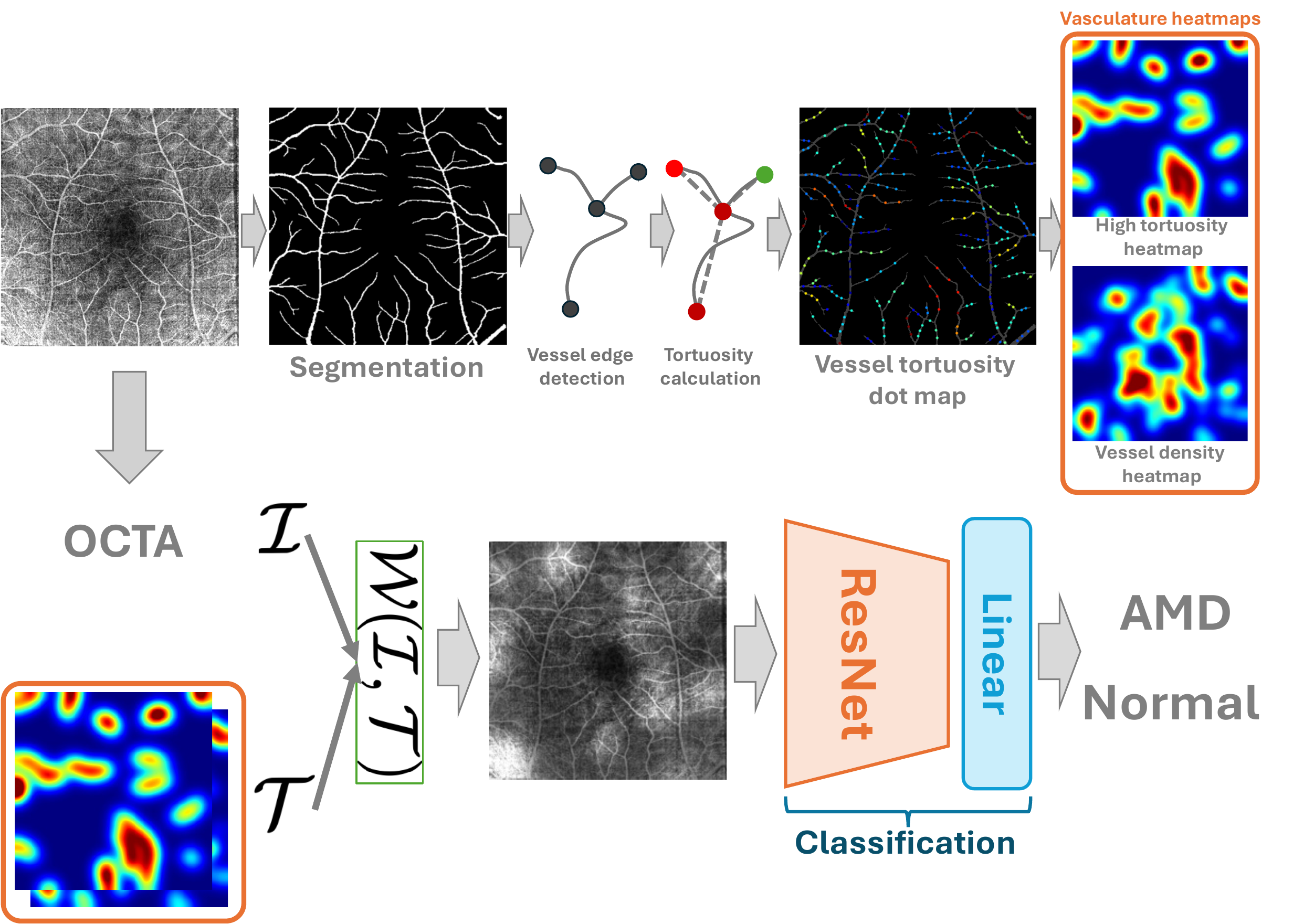}
    \caption{\textbf{Overview of the proposed pipeline} vessel segmentations are used to generate tortuosity or density heatmaps (multi-$\sigma$), which are then separately fused with the full OCTA projection via multiplicative weighting and classified using a ResNet-18 backbone for AMD detection.}
    \label{fig:architecture}
\end{figure}
We design a novel framework that leverages tortuosity and density measurements of the inner retinal vessels as captured on OCTA retinal vessel masks for the arteries, veins, and capillaries, to produce separate tortuosity and density heatmaps used as forms of external attention to a CNN (Fig.~\ref{fig:architecture}). 
We compute vessel-type--specific tortuosity and dropout heatmaps from artery, vein, and capillary masks and fuse each map with the OCTA projection via per-pixel multiplicative weighting prior to ResNet-18 classification (Fig.~\ref{fig:architecture}).
%
\subsection{Tortuosity Heatmap Computation}
\begin{figure}
    \centering
    \includegraphics[width=\linewidth]{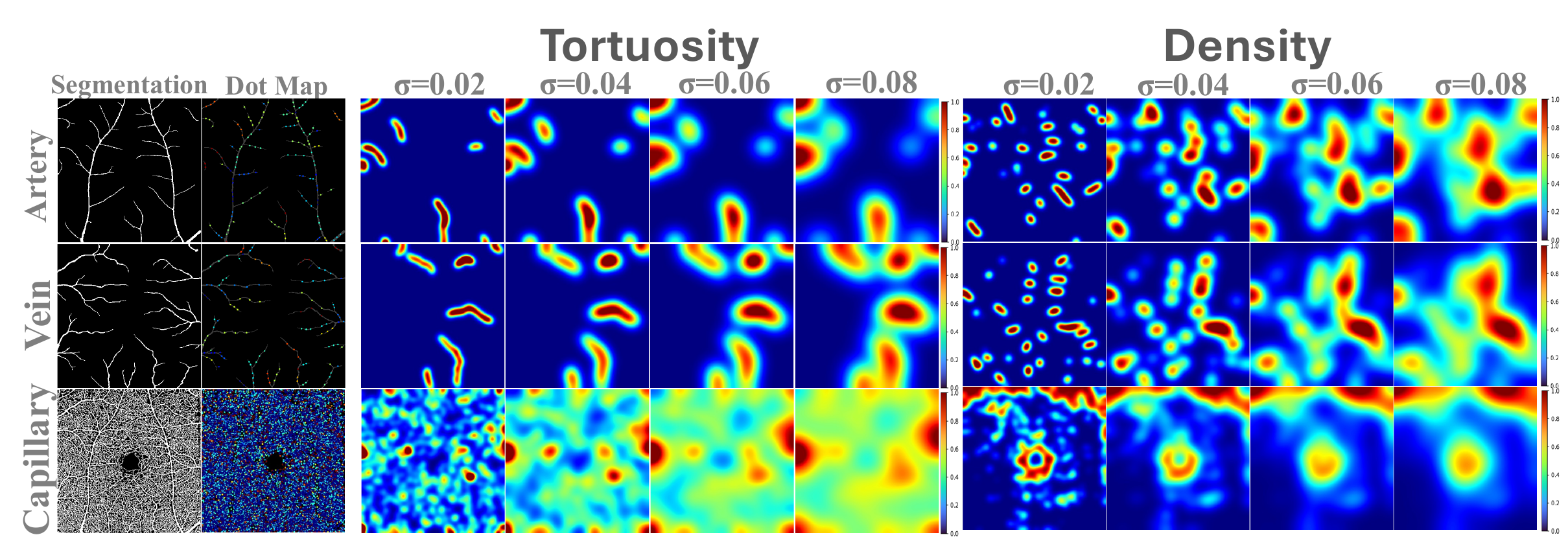}
    \caption{\textbf{Attention heatmaps.} Segmentation labels, tortuosity dot maps showing relative tortuosity of each segment, high tortuosity (red hotspots indicate high tortuosity) and low vessel density heatmaps (red hotspots indicate low density).}
    \label{fig:maps}
\end{figure}

For each 
OCTA, the binary vessel masks corresponding to the arteries, veins and capillaries of the inner retina were used to generate tortuosity heatmaps highlighting regions with highly tortuous vessels. 
Graph-based extraction of vessel edges was performed, such that bifurcation points were treated as nodes, and vessel segments between bifurcation points were treated as edges. 
Endpoints were defined as any skeleton pixels with only 1 neighboring pixel. The total curved length of a vessel edge was defined as: 
$
    L_i = \sum_{j=1}^{n_i-1} \| P_i(j+1) - P_i(j) \|_2$,
where $L_i$ is the total curve length of the edge in pixels, $P_i$ is the ordered list of pixel coordinates for edge $i$, $n_i$ is the number of pixels comprising edge $i$, and $P_i(j)$ is the position of the $jth$ pixel. The total chord length was defined as: 
$C_i = \left\| P_i(n_i) - P_i(1) \right\|_2$, where $C_i$ refers to the straight-line, shortest distance between the first and last pixel of edge $i$, $P_i(1)$ refers to the starting pixel of the edge, and $P_i(n_i)$ refers to the ending pixel of the edge. Based on Hart et al.~\cite{hart1999measurement}, tortuosity for the edge was therefore defined as: $
    T_i = \frac{L_i}{C_i}$.
To create tortuosity heatmaps that would focus on highly tortuous vessels, we selected segments above the 85th percentile of the distribution of $T_i^{\text{excess}}$ in the whole mask. We used the 85th percentile to focus the prior on the most tortuous segments while keeping enough vessel segments for stable smoothing across images. Each selected edge was assigned a weight 
$w_i = (N_i)^{\alpha}(T_i^{\text{excess}})^{\beta}$,
where $N_i$ is the pixel count of the edge, $\alpha$ controls dependence on length of the vessel segment, and $\beta$ controls dependence on curvature magnitude of the vessel segment. This weighting allowed for increasing the influence of longer and highly tortuous segments within the tortuosity spatial maps. After identifying the subset of segments with high tortuosity, we constructed an impulse map used to localize tortuosity hotspots. We began by creating an empty image, and for each selected tortuous segment $i$, we computed the total segment weight $w_i$ and distributed this weight uniformly across all pixels comprising that segment, so each segment would contribute exactly $w_i$, resulting in a global map containing high-frequency impulses located at the most tortuous vessel segment locations. To obtain a continuous spatial distribution of tortuosity, we created multiscale Gaussian heatmaps by convolving the impulse map with Gaussian kernels of increasing width (0.02, 0.04, 0.06, 0.08). These four scales allowed us to capture and assess both local and region-level remodeling. Four Gaussian smoothing scales were used in accordance with the equation
$\sigma_j = f_j \cdot M$
where $\sigma_j$ is the standard deviation of the Gaussian at scale $j$, $f_j$ is the scaling factor (0.02, 0.04, 0.06, or 0.08), and $M$ is the largest image dimension (height or width). The 2D Gaussian kernel used for smoothing was: 
\[
G_{\sigma_j}(u,v)=\frac{1}{2\pi \sigma_j^2}\exp\left(-\frac{u^2+v^2}{2\sigma_j^2}\right)
\]
where $G_{\sigma_j}$ is the Gaussian weighting function, $\sigma_j$ is the smoothing scale, $u$ is the horizontal offset in pixels from the center of the kernel, and $v$ is the vertical offset. Each $f_j$ produces a different smoothed tortuosity heatmap: $D_{\sigma_j}(x,y) = (G_{\sigma_j} * I)(x,y)$,
where $*$ means convolution and $I$ provides impulses from the tortuous vessel segments. Larger sigma values produce more diffuse hotspots, while smaller values preserve localized detail. 
Capillary segmentations were binarized using Otsu’s global thresholding, and connected components smaller than 5 pixels were removed. Skeletonization was performed via morphological thinning to preserve the capillary structure. All other components of the pipeline were identical. Thus, four heatmaps (one per $\sigma$) were generated for arteries, veins and capillaries separately (Fig.~\ref{fig:maps}).


\begin{table*}[t]
\centering
\scriptsize
\caption{Baseline performance without attention. Values are mean $\pm$ standard deviation.}
\label{tab:no_attention}
\begin{tabular}{lcccc}
\hline
\textbf{Method} & \textbf{Bal Acc} & \textbf{Spec} & \textbf{Sens} & \textbf{AUC} \\
\hline
No Attention & 0.86$\pm$0.07 & 0.87$\pm$0.12 & 0.86$\pm$0.12 & 0.90$\pm$0.08 \\
\hline
\end{tabular}
\end{table*}

\begin{table*}[t]
\centering
\scriptsize
\caption{Performance comparison across external attention types, vessel categories, and smoothing levels. Values are mean $\pm$ standard deviation.}
\label{tab:results}
\resizebox{0.7\textwidth}{!}{
\begin{tabular}{c c|cccc|cccc}
\hline
& & \multicolumn{4}{c|}{\textbf{High Tortuosity External Attention}} & \multicolumn{4}{c}{\textbf{Low Density External Attention}} \\
\hline
\textbf{Vessel} & $\sigma$ &
\textbf{Bal Acc} & \textbf{Spec} & \textbf{Sens} & \textbf{AUC} &
\textbf{Bal Acc} & \textbf{Spec} & \textbf{Sens} & \textbf{AUC} \\
\hline
\hline

\multirow{4}{*}{\rotatebox{90}{Artery}}
& 0.02 & 0.83$\pm$0.06 & 0.84$\pm$0.04 & 0.82$\pm$0.15 & 0.87$\pm$0.09  
         & 0.76$\pm$0.13 & 0.69$\pm$0.21 & 0.83$\pm$0.13 & 0.76$\pm$0.12 \\
& 0.04 & 0.79$\pm$0.07 & 0.81$\pm$0.07 & 0.77$\pm$0.16 & 0.83$\pm$0.11  
         & 0.78$\pm$0.04 & 0.85$\pm$0.10 & 0.72$\pm$0.13 & 0.81$\pm$0.06 \\
& 0.06 & 0.83$\pm$0.07 & 0.81$\pm$0.12 & 0.84$\pm$0.15 & 0.85$\pm$0.11  
         & 0.74$\pm$0.02 & 0.89$\pm$0.08 & 0.58$\pm$0.09 & 0.78$\pm$0.05 \\
& 0.08 & 0.85$\pm$0.10 & 0.81$\pm$0.16 & 0.89$\pm$0.12 & 0.89$\pm$0.07  
         & 0.72$\pm$0.05 & 0.89$\pm$0.09 & 0.56$\pm$0.08 & 0.71$\pm$0.06 \\
\hline

\multirow{4}{*}{\rotatebox{90}{Vein}}
& 0.02 & 0.83$\pm$0.08 & 0.78$\pm$0.10 & 0.88$\pm$0.08 & 0.83$\pm$0.07  
         & 0.74$\pm$0.08 & 0.73$\pm$0.18 & 0.75$\pm$0.13 & 0.76$\pm$0.10 \\
& 0.04 & 0.83$\pm$0.05 & 0.86$\pm$0.06 & 0.79$\pm$0.10 & 0.84$\pm$0.06  
         & 0.75$\pm$0.10 & 0.83$\pm$0.14 & 0.67$\pm$0.20 & 0.78$\pm$0.11 \\
& 0.06 & 0.79$\pm$0.10 & 0.80$\pm$0.17 & 0.77$\pm$0.17 & 0.80$\pm$0.10  
         & 0.74$\pm$0.09 & 0.84$\pm$0.13 & 0.65$\pm$0.16 & 0.82$\pm$0.08 \\
& 0.08 & 0.80$\pm$0.06 & 0.81$\pm$0.14 & 0.79$\pm$0.13 & 0.83$\pm$0.09  
         & 0.81$\pm$0.07 & 0.82$\pm$0.15 & 0.79$\pm$0.13 & 0.81$\pm$0.10 \\
\hline

\multirow{4}{*}{\rotatebox{90}{Capillary}}
& 0.02 & 0.76$\pm$0.09 & 0.87$\pm$0.08 & 0.65$\pm$0.20 & 0.77$\pm$0.11  
         & 0.80$\pm$0.07 & 0.80$\pm$0.08 & 0.79$\pm$0.11 & 0.82$\pm$0.06 \\
& 0.04 & 0.80$\pm$0.09 & 0.77$\pm$0.09 & 0.84$\pm$0.11 & 0.81$\pm$0.10  
         & 0.84$\pm$0.08 & 0.79$\pm$0.15 & 0.88$\pm$0.11 & 0.86$\pm$0.08 \\
& 0.06 & 0.80$\pm$0.07 & 0.84$\pm$0.10 & 0.77$\pm$0.07 & 0.83$\pm$0.07  
         & 0.86$\pm$0.07 & 0.78$\pm$0.15 & 0.93$\pm$0.09 & 0.87$\pm$0.07 \\
& 0.08 & 0.78$\pm$0.06 & 0.84$\pm$0.11 & 0.72$\pm$0.12 & 0.80$\pm$0.06  
         & 0.87$\pm$0.05 & 0.88$\pm$0.07 & 0.86$\pm$0.11 & 0.88$\pm$0.07 \\
\hline

\end{tabular}
}
\end{table*}
\subsection{Vessel Density Computation}
For each OCTA, we generated three sets of attention maps emphasizing regions with locally sparse vasculature, using the artery, vein, and capillary binary masks separately; the motivation stems from the clinical implication of vasculature dropout in AMD pathogenesis. Otsu thresholding was first used to return a binary mask. Local vessel density was measured by convolving the binary vessel mask with a circular disk kernel of radius 10 pixels for capillaries, $
D(x,y) = M * K_r(x,y)$
where $M$ is the binary mask, $D(x,y)$ is the number of vessel pixels within the radius $r$ centered at $(x,y)$, and $K_r$ is the disk kernel. We set $r=10$ to represent a local neighborhood large enough to capture meaningful microvascular rarefaction yet small enough to retain spatial heterogeneity. After computing $D(x,y)$, the density map was normalized to $[0,1]$. We defined a sparsity map: $
S(x,y) = 1 - D_{\text{norm}}(x,y)$
where $D_{\text{norm}}(x,y)$ refers to the normalized local density. Thus, $S(x,y)$ increases as vessel density decreases.\\
An impulse map $I(x,y)$ was then generated, assigning nonzero weights only to the pixels whose sparsity exceeds the sparsity threshold. The map was constructed as:  
\[
I(x,y) = 
\begin{cases}
S(x,y), & \text{if } M(x,y)=1 \text{ and } S(x,y)\ge T, \\
0, & \text{otherwise},
\end{cases}
\]
where $I(x,y)$ contains nonzero values only at vessel centerline locations lying within locally sparse neighborhoods. To generate smooth multi-scale low-density maps, each impulse image was blurred with Gaussian kernels at four spatial scales. For an image of size $H \times W$, the Gaussian standard deviation was defined as $
\sigma = f \cdot \max(H,W)$,
where $f$ was either 0.02, 0.04, 0.06, or 0.08 (Fig.\ 2). Each blurred map was then normalized using its 99th percentile for use as an external soft attention map in a CNN.


\subsection{Classification of AMD vs Normal Eyes}

\textbf{OCTA-only model.} We trained and evaluated a binary classifier to distinguish AMD eyes from normal eyes using the OCTA-500 dataset~\cite{li2024octa}. Model performance was estimated with a 5-fold cross validation. OCTA projection images were read as single-channel grayscale and resized to $224\times224$ pixels. Intensities were normalized using ImageNet-1K statistics, adapted from 3 channels to 1 channel. The backbone of the CNN was a ResNet-18~\cite{he2016deep} initialized with ImageNet-1K weights. Because the input was grayscale, the first ConvLayer layer was replaced with a 1-channel layer whose weights were determined by averaging the pretrained RGB filters across channels. The final fully connected layer was replaced with a single-neuron head producing one logit, giving a binary output of AMD or normal.
\subsection{External Multiplicative Attention from Tortuosity and Density Heatmaps}
To assess how tortuosity or density information, as represented in the high-tortuosity and low-vessel-density heatmaps, would affect classification of AMD versus normal eyes and provide interpretability, we incorporated these priors via per-pixel multiplicative attention into separate classifiers with the same architecture as the full-OCTA baseline.\\
Each grayscale heatmap was first normalized over only the nonzero regions. The normalized heatmap was mapped to a multiplicative weight:
\[
\mathcal{W}(x,y) = \text{ATTEN}_{\min} + A(x,y)\left(\text{ATTEN}_{\max}-\text{ATTEN}_{\min}\right)
\]
where $\text{ATTEN}_{\min} = 0.5$ and $\text{ATTEN}_{\max} = 1.5$. The final modulated image was $
R_{\text{fused}}(x,y) = R(x,y)\cdot \mathcal{W}(x,y)$,
where $R(x,y)$ is the original OCTA image. Bounding attention to [0.5, 1.5] allowed for suppressing regions of low tortuosity or high density while amplifying regions of high tortuosity or low density, excluding extreme intensity scaling that would destabilize training.
\section{Experiments and Results}
\subsection{Experiments}
\textbf{Datasets.} We used OCTA imaging and retinal vessel segmentation labels from the OCTA-500 dataset~\cite{li2024octa}, including 43 eyes with age-related macular degeneration (AMD) and 91 healthy eyes, all acquired using a 6×6 mm² field of view on a 70 kHz spectral-domain OCT system (RTVue-XR, Optovue; 840 nm). One random eye per subject was imaged at Jiangsu Province Hospital between March 2018 and July 2020. OCTA full-projection images were generated from 3D OCTA volumes using the split-spectrum amplitude-decorrelation angiography (SSADA) algorithm~\cite{jia2012split}.\\
For each eye, the dataset provides a full-projection OCTA image along with multilabel segmentations of arteries, veins, capillaries, and the foveal avascular zone (FAZ). Vessel labels correspond to the maximum projection of the inner retina (between the inner limiting membrane and outer plexiform layer), enabling accurate delineation of inner-retinal vasculature and FAZ geometry.\\
\textbf{Implementation details.} Training followed a two-stage schedule. During the warmup stage, the backbone was frozen and the classification head only was optimized for $W = 2$ epochs, with a learning rate of $10^{-3}$. During the fine-tuning stage, all parameters were unfrozen and optimized with a learning rate of $10^{-4}$. Both stages used the Adam optimizer with weight decay $10^{-4}$, batch size 16, and a maximum of 20 epochs per fold. A ReduceLROnPlateau scheduler monitored validation balanced accuracy and reduced the learning rate by a factor of 0.5 after two stagnant epochs. CUDA/cuDNN benchmarking was enabled. We used binary focal loss with logits:$\text{FocalLoss}(y,\hat{p}) = -\alpha_t(1-p_t)^{\gamma}\log(p_t)$
where $\hat{p}=\sigma(z)$ is the sigmoid probability of AMD, $p_t=\hat{p}$ if $y=1$ and $p_t = 1-\hat{p}$ if $y=0$. The class weight was $\alpha_t = 0.75$ for AMD samples and $1-\alpha_t$ for normal samples, and the focusing parameter $\gamma=2.0$. A 5-fold stratified cross validation was used, and for each fold the model with the highest validation balanced accuracy was selected. Balanced accuracy, sensitivity, specificity, and AUROC were reported.
\begin{figure}[t]
    \centering
    \includegraphics[width=\linewidth]{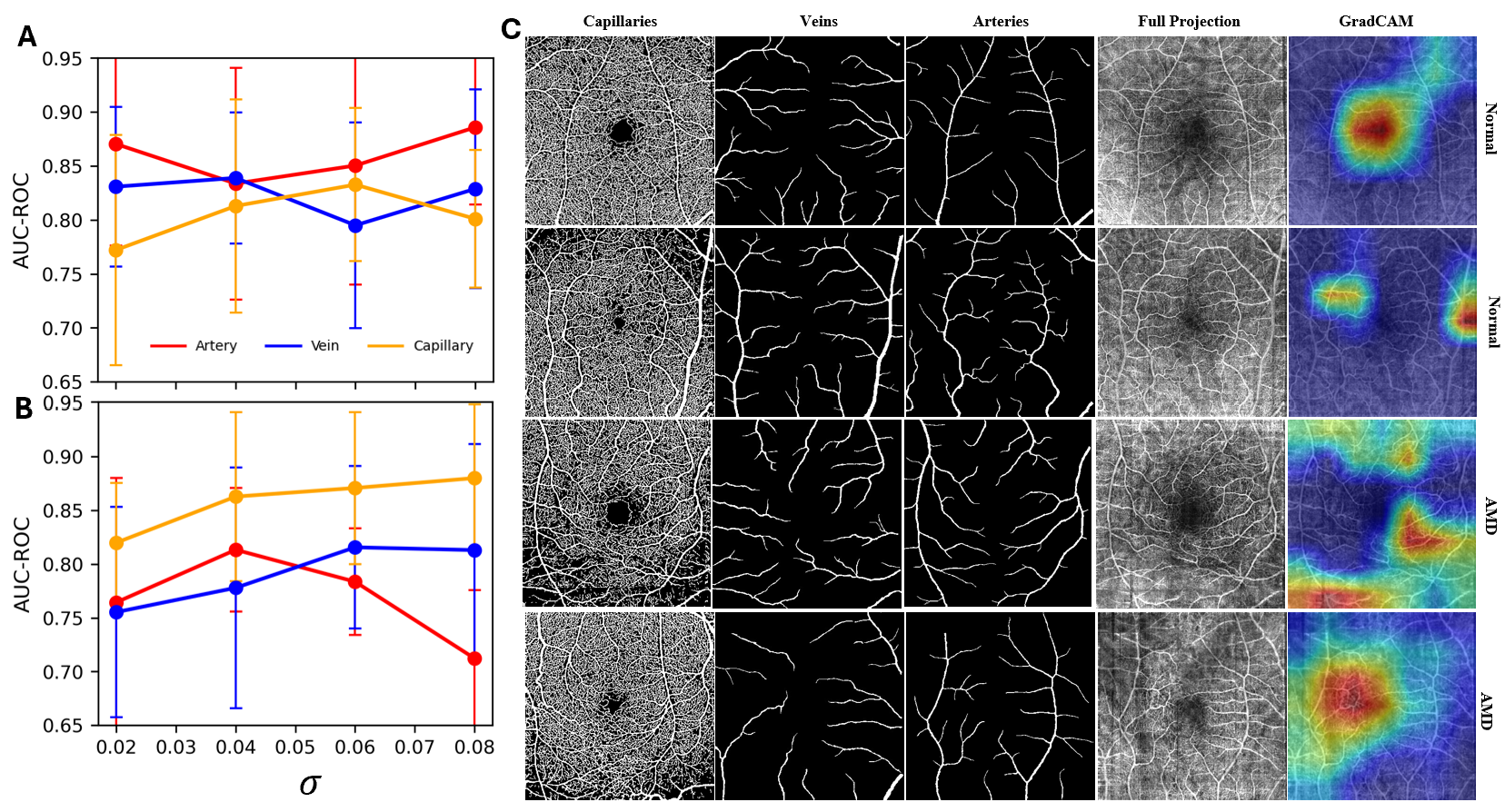}
    \caption{AUC-ROC vs. $\sigma$ for vessel tortuosity (\textbf{A}) and density (\textbf{B}) heatmap-attentions are shown. GradCAM results are shown in \textbf{C}.}
    \label{fig:results}
\end{figure}
\subsection{Results and Discussion}
Tab.~\ref{tab:no_attention} reports the OCTA-only baseline results (no attention), while Tab~\ref{tab:results} reports results for all binary classifications using external multiplicative attention from tortuosity or vessel sparsity maps. These results demonstrate the effects of adding vessel priors to the DL pipeline. Attention variants show similar mean AUC to the baseline model, indicating that performance is comoparable to baseline, with the addition of biomarker-guided interpretability.
Fig.~\ref{fig:results}A and B plot AUC versus the smoothing parameter $\sigma$. The arterial tortuosity heatmap yielded the strongest and most stable performance among tortuosity-based attention maps, with accuracy and AUC largely improving as the smoothing parameter $\sigma$ increased. Clinically, this trend is consistent with early AMD pathology: retinal arterioles exhibit stiffening and impaired auto-regulation, leading to subtle, segment-level shape deviations. Larger $\sigma$ values highlight these broad, spatially coherent tortuosity patterns while suppressing pixel-scale noise. Because arterioles are tightly coupled to choroidal perfusion—thought to deteriorate early in AMD—the enhanced performance of smoothed arterial tortuosity maps supports the role of arteriolar remodeling as a meaningful vascular biomarker of AMD. The vein heatmap dropped in performance at intermediate values. The capillary heatmap performed best at mid-range $\sigma$ and benefited overall from smoothing.\\
Capillary-derived maps showed two consistent trends: tortuosity-based maps improved with moderate smoothing, while low-density capillary maps outperformed all other density-based attention maps, especially at larger $\sigma$ values. Clinically, this aligns with the fact that capillary rarefaction—manifested as parafoveal capillary loss, FAZ enlargement, and localized nonperfusion—is a hallmark of early AMD. Moderate-to-large $\sigma$ emphasizes these broader, spatially diffuse ischemic patterns rather than isolated capillary gaps, making the smoothed capillary density maps particularly effective. This supports extensive evidence that capillary perfusion loss is among the earliest and most robust vascular biomarkers of AMD. Our results reinforce the established finding that capillary rarefaction is an important AMD biomarker.\
As $\sigma$ increased, the vessel-dropout maps progressively highlighted more of the FAZ, which is clinically consistent with AMD-related FAZ enlargement. Larger  $\sigma$ values smooth the FAZ boundary and capture region-level deviations—such as irregularity, asymmetry, and expansion—that are well-documented OCTA biomarkers correlating with visual dysfunction and disease severity. This enhanced emphasis on FAZ morphology likely contributed to the improved performance of the capillary density–based models.\\
GradCAM visualizations (Fig.~\ref{fig:results}C) indicate that the CNN focuses on regions of reduced capillary density, outward displacement of capillary terminals near the FAZ, and localized arteriolar tortuosity. These correspond to well-established AMD features—including parafoveal nonperfusion, capillary rarefaction, FAZ shape irregularity, and arteriolar dysregulation—suggesting that the model is leveraging clinically plausible vascular cues associated with AMD pathology.

\section{Conclusion}

We presented a biomarker-guided attention framework that integrates vessel tortuosity and vasculature dropout maps into OCTA-based AMD classification. The observed performance trends, particularly the importance of arterial tortuosity and capillary dropout, closely mirror known microvascular changes in early AMD. Our findings highlight the potential of vascular biomarker priors to enhance interpretability and clinical alignment in DL models for AMD detection. Future work will extend these findings across AMD disease stages.

\section{Compliance with Ethical Standards}
This study, conducted on open-source data, did not require ethical approval.

\section{Acknowledgements}
This research was partially supported by National Institutes of Health (NIH) and
National Cancer Institute (NCI) grants 1R21CA25849301A1, 1R01CA297843-01, 3R21CA258493-02S1,
1R03DE033489-01A1, and National Science Foundation (NSF) grant 2442053. The content is solely the
responsibility of the authors and does not necessarily
represent the official views of the National Institutes of
Health.

\bibliographystyle{IEEEbib}
\bibliography{strings,refs}

\end{document}